\documentclass[10pt,twocolumn,letterpaper]{article}

\usepackage{iccv}
\usepackage{times}
\usepackage{epsfig}
\usepackage{graphicx}
\usepackage{amsmath}
\usepackage{amssymb}

\usepackage{subfloat}
\usepackage{booktabs}
\usepackage{amsfonts}
\usepackage{bm}
\usepackage{multicol}
\usepackage{multirow}
\usepackage{txfonts}
\usepackage{pifont}
\usepackage{makecell}
\usepackage{enumitem}
\usepackage{threeparttable}
\RequirePackage[format=plain,labelformat=simple,labelsep=period,font=small,compatibility=false]{caption}
\RequirePackage[font=footnotesize,skip=3pt,subrefformat=parens]{subcaption}

\usepackage[pagebackref=true,breaklinks=true,letterpaper=true,colorlinks,bookmarks=false]{hyperref}

\iccvfinalcopy 

\def\iccvPaperID{5120} 

\ificcvfinal\fi

\begin{document}

\title{Learning Global-aware Kernel for Image Harmonization}
\author{Xintian Shen$^1$\thanks{Equal contribution.}
~ ~Jiangning Zhang$^2$\footnotemark[1]
~ ~Jun Chen$^1$
~ ~Shipeng Bai$^1$ \\
~ ~Yue Han$^1$
~ ~Yabiao Wang$^2$
~ ~Chengjie Wang$^{2,3}$
~ ~Yong Liu$^1$\thanks{Corresponding author.} \\
\normalsize $^1$ APRIL Lab, Zhejiang University ~ ~ $^2$Youtu Lab, Tencent ~ ~ $^3$Shanghai Jiao Tong University \\
{\tt\small [22132133, 186368, junc, shipengbai, 
22132041]@zju.edu.cn,} \\
{\tt\small[caseywang, jasoncjwang]@tencent.com, yongliu@iipc.zju.edu.cn}
}

\maketitle
\ificcvfinal\thispagestyle{empty}\fi

\begin{abstract}

Image harmonization aims to solve the visual inconsistency problem in composited images by adaptively adjusting the foreground pixels with the background as references. Existing methods employ local color transformation or region matching between foreground and background, which neglects powerful proximity prior and independently distinguishes fore-/back-ground as a whole part for harmonization. As a result, they still show a limited performance across varied foreground objects and scenes. To address this issue, we propose a novel Global-aware Kernel Network (GKNet) to harmonize local regions with comprehensive consideration of long-distance background references.
Specifically, GKNet includes two parts, \ie, harmony kernel prediction and harmony kernel modulation branches. The former includes a Long-distance Reference Extractor (LRE) to obtain long-distance context and Kernel Prediction Blocks (KPB) to predict multi-level harmony kernels by fusing global information with local features. To achieve this goal, a novel Selective Correlation Fusion (SCF) module is proposed to better select relevant long-distance background references for local harmonization. The latter employs the predicted kernels to harmonize foreground regions with local and global awareness. Abundant experiments demonstrate the superiority of our method for image harmonization over state-of-the-art methods, \eg, achieving 39.53dB PSNR that surpasses the best counterpart by +0.78dB $\uparrow$; decreasing fMSE/MSE by 11.5\%$\downarrow$/6.7\%$\downarrow$ compared with the SoTA method. Code will be available at \href{https://github.com/XintianShen/GKNet}{here}.

\end{abstract}

\section{Introduction}

\label{sec:intro}
Image composition aims to synthesize foreground objects from one image into another, which is a common task in image editing. However, human eyes could clearly distinguish synthetic images due to the visual inconsistency between foreground and background in composited images.
In attempting to solve the photo-unrealistic problem, image harmonization is proposed to adjust the foreground objects based on the illumination and color tone in background environment, which plays an important role in image editing.

\begin{figure}[!t]
    \centering
    \begin{subfigure}{0.32\linewidth}
    \includegraphics[width=0.95\linewidth]{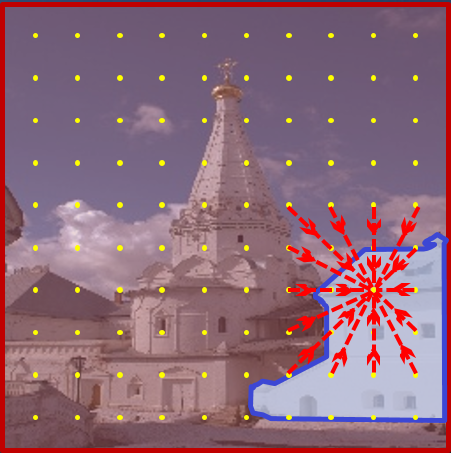}
    \caption{Local-translation}
    \label{fig:pixel-level}
    \end{subfigure}
    \begin{subfigure}{0.32\linewidth}
    \includegraphics[width=0.95\linewidth]{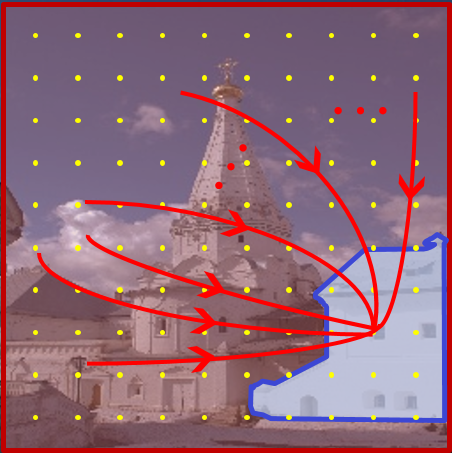}
    \caption{Region-matching}
    \label{fig:image-level}
    \end{subfigure}
    \begin{subfigure}{0.32\linewidth}
    \includegraphics[width=0.95\linewidth]{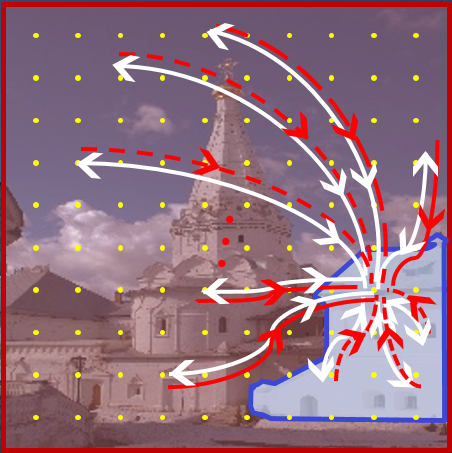}
    \caption{Ours}
    \label{fig:intro_ours}
    \end{subfigure}
    \caption{Comparison of background reference methods in harmonization. \textcolor{blue}{Blue}/\textcolor{red}{Red} region represent foreground/background, respectively, and white/red arrows refer to interaction/injection, respectively. (a) Local-translation methods reference nearby pixels. (b) Region-matching methods transfer reference with a unified view of fore-/back-ground region. (c) Our method interacts long-distance reference and injects it with short-distance consideration. 
 }
    \label{fig:intro}
    \vspace{-1em}
\end{figure}

\begin{figure*}[t]
  \centering
  \captionsetup[subfloat]{labelformat=empty}
  \begin{subfigure}{0.67\linewidth}
    \resizebox{\linewidth}{!}{
    \subfloat[Mask]{
    \begin{minipage}[b]{0.2\linewidth}
    \includegraphics[width=1\linewidth]{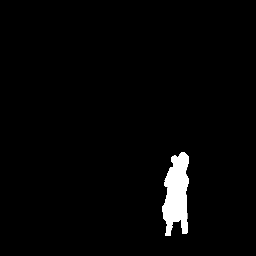}\vspace{1pt}
    \includegraphics[width=1\linewidth]{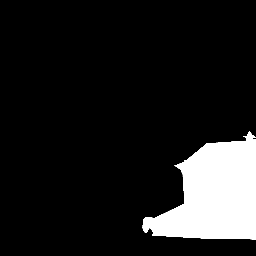}\vspace{1pt}
    \end{minipage}}
    \subfloat[Input]{
    \begin{minipage}[b]{0.2\linewidth}
    \includegraphics[width=1\linewidth]{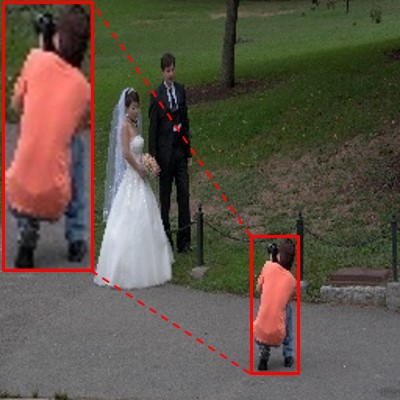}\vspace{1pt}
    \includegraphics[width=1\linewidth]{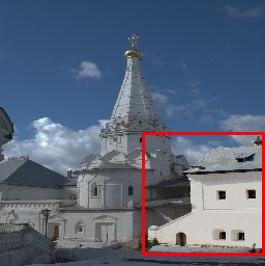}\vspace{1pt}
    \end{minipage}}
    \subfloat[RainNet\cite{Cong_2020_CVPR}]{
    \begin{minipage}[b]{0.2\linewidth}
    \includegraphics[width=1\linewidth]{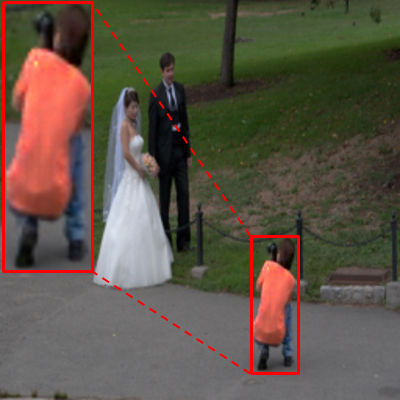}\vspace{1pt}
    \includegraphics[width=1\linewidth]{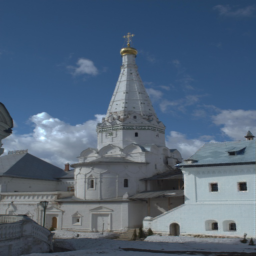}\vspace{1pt}
    \end{minipage}}
    \subfloat[iS$^2$AM\cite{Sofiiuk_2021_WACV}]{
    \begin{minipage}[b]{0.2\linewidth}
    \includegraphics[width=1\linewidth]{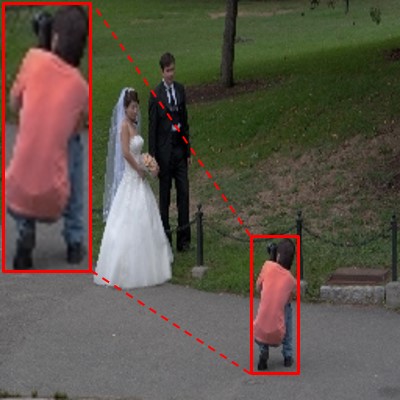}\vspace{1pt}
    \includegraphics[width=1\linewidth]{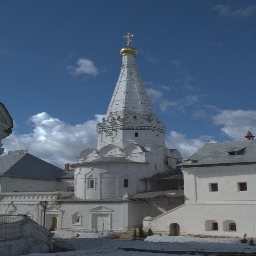}\vspace{1pt}
    \end{minipage}}
    \subfloat[Ours]{
    \begin{minipage}[b]{0.2\linewidth}
    \includegraphics[width=1\linewidth]{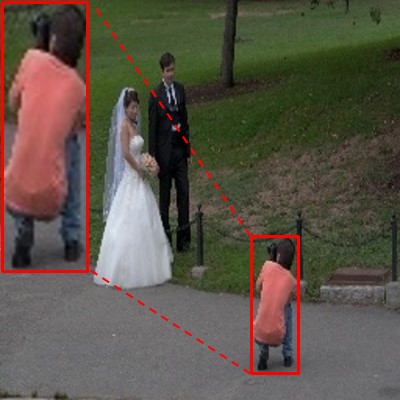}\vspace{1pt}
    \includegraphics[width=1\linewidth]{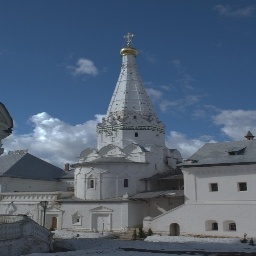}\vspace{1pt}
    \end{minipage}}
    \subfloat[Real(GT)]{
    \begin{minipage}[b]{0.2\linewidth}
    \includegraphics[width=1\linewidth]{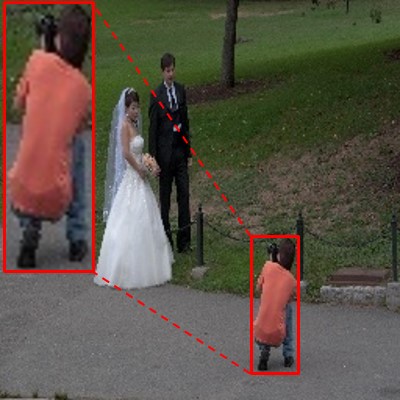}\vspace{1pt}
    \includegraphics[width=1\linewidth]{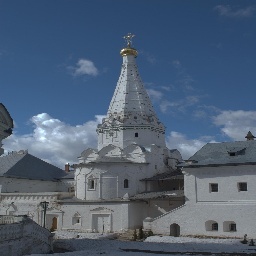}\vspace{1pt}
    \end{minipage}}}
    \setcounter{subfigure}{0}
    \label{fig:intro-example}
  \end{subfigure}
  \hspace{1pt}
  \begin{subfigure}{0.29\linewidth}
    \includegraphics[width=\linewidth]{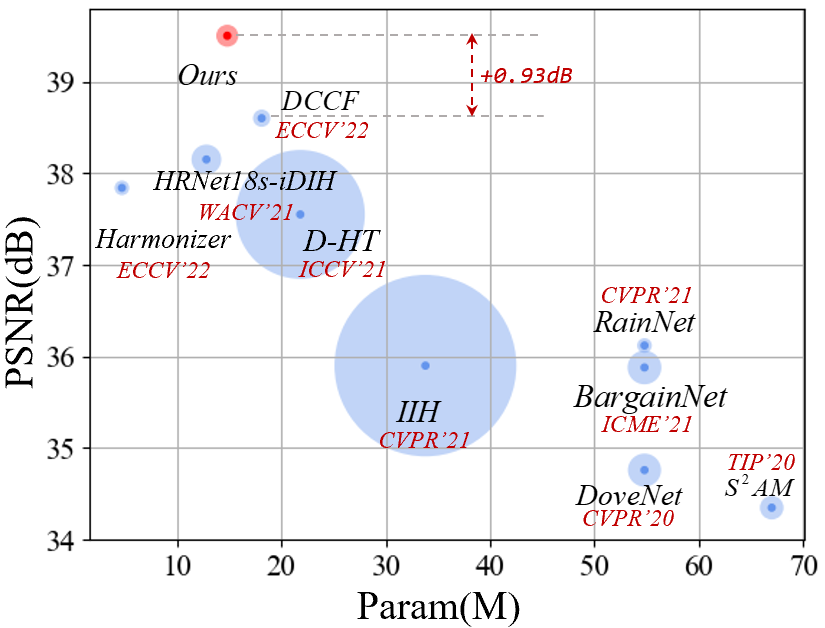}
    \label{fig:exampe_intro}
  \end{subfigure}
  \caption{\textbf{Left:} Two challenging samples in image harmonization. Mask in column one and Red boxes represents the foreground. \textbf{Right:} Performance comparison with SOTA methods in terms of PSNR and model size. The circle size represents the floating-point number.}
  \label{fig:intro_2}
\end{figure*}

Traditional image harmonization approaches are mainly based on low-level feature matching, which is only effective for specific scenes. Recently, numerous learning-based methods have achieved remarkable progress by addressing image harmonization as a generation task. Existing learning-based methods could be categorized from two angles, \ie, \textit{local-translation} and \textit{region-matching}. 1) The former employs a convolutional encoder-decoder to learn a foreground pixel-to-pixel translation~\cite{Tsai_2017_CVPR,cun_2020_TIP}. But a shallow CNN only captures limited surrounding background. As shown in Figure~\ref{fig:pixel-level}, these approaches harmonize the current pixel with local references, which is insufficient for harmonization as inner foreground pixels could not attach background reference. Besides, related long-distance references are effective in some cases. 2) The latter region matching methods~\cite{Ling_2021_CVPR,Cong_2020_CVPR} distinguish foreground and background regions as two styles or domains. As shown in Figure~\ref{fig:image-level}, they tackle harmonization as a matching problem with a unified view of these two regions by statistics components or discriminators. Though these approaches harmonize images with a broader reference range, they totally neglect the spatial variations in two regions. Hang \etal~\cite{Hang_2022_CVPR} begin to notice this problem and add attention-based references in region matching method~\cite{Ling_2021_CVPR}. 
But they still separate two regions independently and harmonize foreground by unified matching without considering foreground spatial variations.

To further illustrate existing problems, we provide two common harmonization cases in Figure~\ref{fig:intro_2}. In the first case, a small foreground object appears in the background with obvious color changes. The region-matching method RainNet~\cite{Cong_2020_CVPR} provides a poor color correction result while the local method iS$^2$AM~\cite{Sofiiuk_2021_WACV} could tackle this case well, which indicates that the unified view of background will blend the overall complex color conditions. In the second case, related long-distance references exist in the background, while the local method could only attach insufficient adjacent information. Region-matching method RainNet could obtain whole background blue tone by matching, but it still excessively harmonizes the house due to the unified view. These two cases indicate that local reference is insufficient, but region-matching methods could not model long-distance reference well and will cause unbalanced harmonization problems by rough matching. 

To solve this problem, we rethink essential proximity priors in image harmonization, \ie, when we paste an object into background, the color or light is related to location and will be influenced by its neighboring first. Moreover, the effective long-distance information in background changes with pasted locations, which requires us to learn adaptive references for each part. Inspired by this observation, we propose a novel \emph{Global-aware Kernel Network} (GKNet) to integrate local harmony modulation and long-distance background references, including harmony kernel prediction and harmony kernel modulation. For harmony kernel prediction, we propose a novel global-aware kernel prediction method including \emph{Long-distance Reference Extractor} (LRE) to obtain long-distance references and \emph{Kernel Prediction Blocks} (KPB) to predict multi-level adaptive kernels with selected long-distance references by \emph{Selective Correlation Fusion} (SCF). For kernel modulation, we propose to model local harmony operation by predicted global-aware kernels and multi-level features. Focusing on features in kernel region, kernel modulation is significant in alleviating unbalanced region-matching errors in complex scenes.

To summarize, we make following contributions: 
\begin{itemize}[leftmargin=*]
\vspace{-2pt}
\item With the observation of proximity prior and long-distance references in image harmonization task, we design GKNet to model global-local interaction by learning global-aware harmony kernel, including harmony kernel prediction and harmony kernel modulation.
\vspace{-4pt}
\item For harmony kernel prediction, we propose a kernel prediction branch combined with LRE to model global information and multiple KPB to learn adaptive harmony kernels. For better global references, we design SCF to select relevant long-distance references for local harmonization. For harmony kernel modulation, we propose the method to harmonize local regions in multi-level decoder layers with predicted kernels.
\vspace{-4pt}

\item Extensive experiments demonstrate the superior performance of our methods in both quantitative and qualitative results, noting that our method achieves state-of-the-art results on iHarmony4 datasets.

\end{itemize}

\begin{figure*}[t]
  \centering
  \includegraphics[width=17cm]{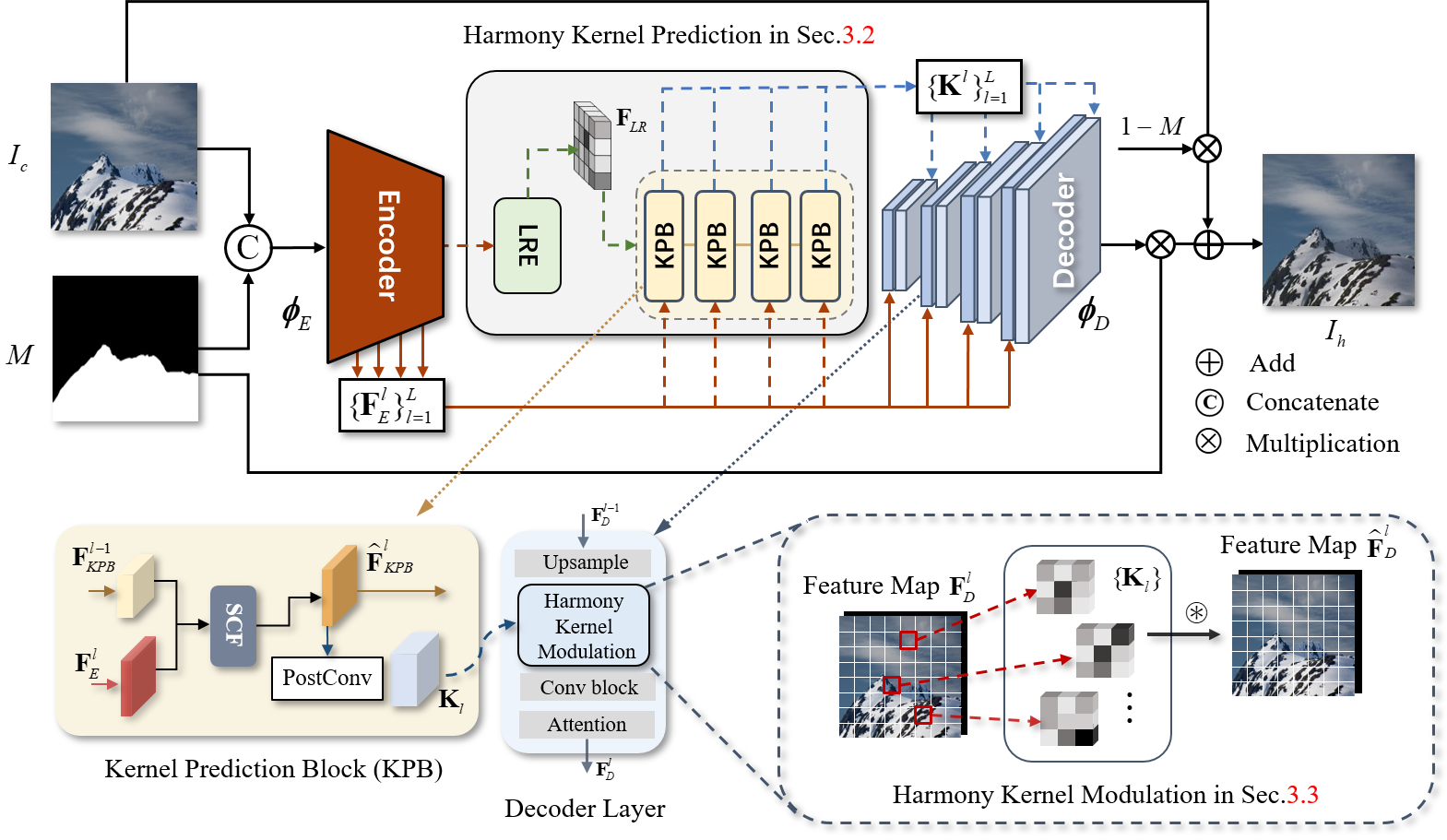}
  \caption{{\bf The overview of our proposed GKNet}, which consists of harmony kernel prediction branch and harmony kernel modulation branch. As shown in gray box, the harmony kernel prediction branch is combined with a \emph{Long-term Reference Extractor} (LRE) and multi-level \emph{Kernel Prediction Blocks} (KPB). As shown in yellow box, we propose \emph{Selective Correlation Fusion} (SCF) module in KPB for better long-distance references. Given a composited image $I_c$ with corresponding foreground mask M, we extract deep features ${\mathbf{F}_E^l}$ from encoder $\boldsymbol\phi_E$. Then, harmony kernel prediction branch utilizes the deepest feature map and $\{\mathbf{F}_E^l\}$ to predict multi-level dynamic harmony kernels $\{\mathbf{K}^l\}$ increasingly. The predicted global-aware kernels are employed for harmony kernel modulation in decoder $\boldsymbol\phi_D$.}
  \label{fig:main_model}
\end{figure*}
\vspace{-5pt}
\section{Related work}

\subsection{Image Harmonization} Traditional image harmonization works have focused on finding a better method for low-level appearances matching between foreground and background regions in images, which includes color statistics~\cite{reinhard2001color,pitie2007linear,xue2012understanding},  gradient information~\cite{jia2006drag,perez2003poisson,tao2010error}, and multi-scale statistical features~\cite{sunkavalli2010multi,lalonde2007using}. However, traditional methods could only be effective in specific scenes. With the advanced generative ability of deep learning, Tsai \etal~\cite{Tsai_2017_CVPR} firstly propose a learning-based encoder-decoder network assisted by a semantic branch. In observation that semantic information is effective in image harmonization, Soffiuk \etal~\cite{Sofiiuk_2021_WACV} also add additional pre-train semantic model to baseline DIH~\cite{Tsai_2017_CVPR} and S$^{2}$AM~\cite{cun_2020_TIP}. Inspired by domain transfer, Cong \etal~\cite{Cong_2020_CVPR} adopt a verification discriminator to distinguish foreground and background domains. Similarly, Ling \etal~\cite{Ling_2021_CVPR} also treat the composited image as two independent parts and apply style transfer idea to match mean-variance statistics. To focus on harmonize foreground region, some methods add attention mechanisms. Cun \etal~\cite{cun_2020_TIP} add a spatial-separated attention module. Guo \etal~\cite{Guo_2021_ICCV} for the first time introduce Transformer architecture to image harmonization. Hang \etal~\cite{Hang_2022_CVPR} add background attention calculation to the style transfer block~\cite{Ling_2021_CVPR}, and they also incorporated the idea of contrast learning. Besides, Guo\etal~\cite{Guo_2021_CVPR} decompose image into reflectance and illumination by autoencoder for separate harmonization based on Retinex theory. Some high-resolution methods~\cite{Harmonizer,xue2022dccf} frame image harmonization as an image-level problem to learn white-box arguments. However, the \emph{above methods neglect spatial proximity prior and could not model long-distance references well}. Instead in this paper, we design a better local modulation method combined with selected long-distance references to alleviate this problem.

\subsection{Dynamic Filtering in Image Editing}
The input-dependent dynamic filtering first proposed by Jia~\cite{jia2016dynamic}~\etal aims to learn position-specific filters on pixel inputs and apply the generated kernels to another input, which has been widely used in numerous vision tasks~\cite{niklaus2017video,jo2018deep,yang2019condconv,dai2017deformable}. This method also shows effectiveness in image editing tasks, such as denoising~\cite{bako2017kernel,mildenhall2018burst,vogels2018denoising}, shadow removing~\cite{fu2021auto}, deraining~\cite{guo2021efficientderain}, image inpainting~\cite{li2022misf}, and blur synthesis~\cite{brooks2019learning}. However, most above methods apply dynamic filtering at image-level filter prediction and utilization. We propose to learn a multi-level global-aware kernel with long-term context references for harmonization.

\subsection{Feature Fusion} 
Feature fusion is to combine features from different layers or branches, which is an omnipresent part of modern neural networks and has been studied extensively. Most previous works~\cite{he2016deep,ronneberger2015u,lin2017feature} for feature fusion focus on the pathways structure design, applying two linear classic methods of summation or concatenation. Recently, benefit from the successful use of Transformer in computer vision~\cite{chen2020generative,dosovitskiy2020image,liu2021swin,zhang2021analogous,zhang2022eatformer,zhang2023rethinking,carion2020end,chen2020generative,esser2021taming,wan2021high,jiang2021transgan,liu2021fuseformer,Guo_2021_ICCV}, some works~\cite{hu2018squeeze,li2019selective,zhang2018exfuse,dai2021attentional,zhang2022resnest,zhang2022scsnet,chen2023better} apply attention mechanism to present nonlinear approaches for feature fusion. As global information is significant in image harmonization, we design a dynamic weighted fusion method to effectively fuse long-distance reference into kernel prediction.

\section{Method}

Given a composited image $I_c$ with its corresponding binary mask M indicating the region to be harmonized, our goal is to learn a network G that outputs harmonized image $I_h$, which could be formulated as $I_h=G(I_c, M)$. To make this composited image $I_c$ look natural, we train our model G to adjust the foreground region $I_f$ in a supervised manner with paired real image $I$. In this paper, we also define the background region as $I_b$, and then the composition process could be formulated as $I_c=I_f \cdot M + (1-M) \cdot I_b$, where $ \cdot $ denotes element-wise multiplication.

\subsection{Overview of Our Network}

As shown in Figure~\ref{fig:main_model}, we design a novel network architecture for image harmonization tasks to allow our network to pay attention to short-distance and long-distance information simultaneously. Following the standard designs in image harmonization works ~\cite{cun_2020_TIP,Sofiiuk_2021_WACV,Ling_2021_CVPR} we use simple U-Net~\cite{ronneberger2015u} with attention blocks~\cite{cun_2020_TIP} as the basic structure. We also take composited RGB image $I_c \in \mathbb{R}^{3 \times H \times W}$ concatenated with foreground region mask M $\in \mathbb{R}^{1 \times H \times W}$ as input.

Motivated by proximity prior in image harmonization, we propose \emph{Global-aware Kernel Network} (GKNet) to learn global-aware harmony kernel for image harmonization, which consists of two branches, harmony kernel prediction and harmony kernel modulation. Firstly, as long-distance reference is crucial for harmonization task, we design global-aware kernel prediction branch to predict harmony kernel with context modeling, which contains a transformer-based \emph{Long-term Reference Extractor} (LRE) to extract global reference and \emph{Kernel Prediction Block} (KPB) to predict harmony kernels. In order to incorporate relevant long-term references for local harmonization, a novel \emph{Selective Correlation Fusion} (SCF) is proposed to select more effective references in backgrounds. Secondly, we design a multi-level harmony kernel modulation in decoder layers to employ the predicted global-aware kernels. The mechanism between global-aware harmony kernel prediction and harmony kernel modulation finally achieves local-global interaction for image harmonization.

\subsection{Harmony Kernel Prediction} 
\label{sec:HKP}

Inspired by recent works in image editing ~\cite{bako2017kernel,guo2021efficientderain,mildenhall2018burst}, we propose to apply dynamic kernels to force the current pixel harmonized with surrounding regions adaptively. This approach effectively makes up for the lack of consideration of proximity prior in previous harmonization works.
However, basic dynamic kernels for image editing tasks such as denoising and deraining are applied with a fixed size at image-level. To predict more proper adaptive kernels for image harmonization, we analyze the following: \textbf{1)} Global modeling is necessary for image harmonization as long-distance references may appear in the background. Hence, we design a novel global-aware harmony kernel prediction branch with LRE to extract global information and KPB to predict global-aware kernels with fusion module SCF. \textbf{2)} Fixed-size kernels applied at image-level could not handle the scale problem well. \eg The pixels inside the large foreground mask can hardly obtain any real background information, while predicting large kernels will bring high computation costs and breaks the intention of proximity prior. Besides, image-level dynamic kernels pay more attention to detailed structure, while in image harmonization we also need to adapt multiple scene variations to harmonize foregrounds at semantic level. In order to adapt to multi-scene and multi-scale problems, we propose to predict kernels in multi-level structures.

~\\
\noindent \textbf{Long-distance Reference Extractor.}
In order to obtain long-term context, we employ l-transformer layers~\cite{dosovitskiy2020image} as our global information extractor. We feed the deepest feature map $\mathbf{F}_E^1$ from CNN encoder into transformer layers. With the down-sampling feature map in low-resolution of $(\frac{w}{r},\frac{h}{r})$, we treat each pixel as a token to generate embeddings. With the multi-head attention mechanism, we obtain global interactive feature $\mathbf{F}_{global} \in \mathbb{R}^{C \times HW}$ after l-transformer layers. After reshaping and post-convolution layer, we obtain the long-term reference feature $\mathbf{F}_{LR}$.

~\\
\noindent {\bf Kernel Prediction Block.} 
To adapt diverse foreground scales and background scenarios, we apply our local operation kernel modulation in multiple decoder levels. Nevertheless, deep-level features contain more semantic information, and shallow features contain more details, we need to predict corresponding adaptive kernels for different level harmony kernel modulation. Thus, our designed global-aware harmony kernel prediction branch is in a multi-level structure to increasingly predict a series of kernels.

In Figure~\ref{fig:main_model}, we show our proposed KPB structure from predicting kernels to harmony kernel modulation operation. The operation in $l$th KPBlock can be formulated as
\begin{equation}
    {\mathbf{K}^l, \mathbf{F}^l_{KPB} = KPBlock (\mathbf{F}^l_E,\mathbf{F}^{l-1}_{KPB})},
\end{equation}
where KPBlock($\cdot$) is the KPBlock to predict $\mathbf{K}^l$. For each KPB, we take $\mathbf{F}^{l-1}_{KPB} \in \mathbb{R}^{C_l \times H_l \times W_l}$ transferred from $(l-1)$th KPB (For the deepest KPB, we input $\mathbf{F}_{LR}$) and the $l$th encoder layer feature $\mathbf{F}^l_E$ as input  (We denote $\mathbf{F}_E^1$ as the deepest layer in encoder), which outputs feature $\mathbf{F}^{l}_{KPB}$ for next KPBlock and the global-aware kernels $\mathbf{K}^l=$ Conv$(\mathbf{F}^{l}_{KPB})$ after post-convolutions.

~\\
\noindent \textbf{Selective Correlation Fusion.}
Long-distance reference $\mathbf{F}_{LR}$ obtained from LRE is then injected into the deepest KPBlock to model global information for local harmonization. The standard way of feature fusion, like concatenation or addition, equally treats low-level and high-level features. To efficiently model long-term information for local harmonization, we propose SCF to select relevant global information by interacting encoder features $\mathbf{F}_E$ and long-distance references based on channel-wise attention mechanism. 

\begin{figure}[!t]
    \centering
    \includegraphics[width=\linewidth]{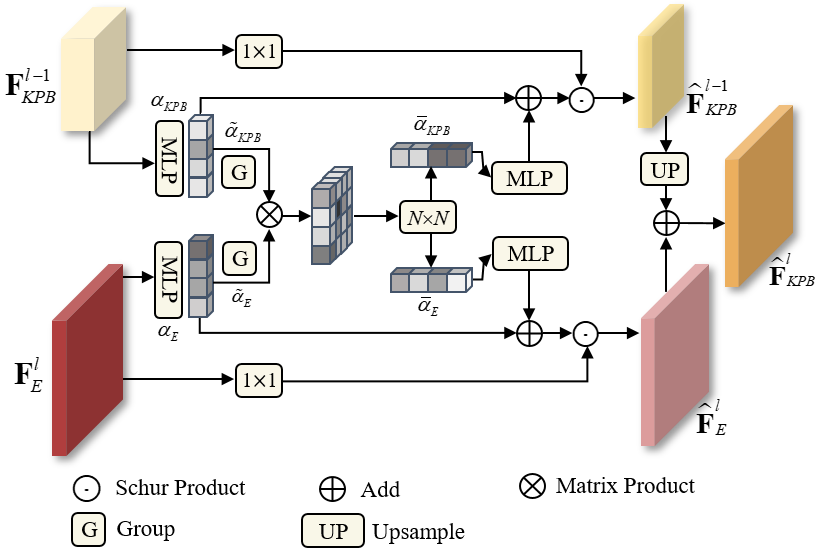}
    \caption{{\bf Schematic diagram of SCF.} The module takes the (l-1)th layer feature $\mathbf{F}^{l-1}_{KPB}$ and encoder feature $\mathbf{F}^{l}_E$ as input and outputs the correlation-aware fusion feature $\hat{\mathbf{F}}_{KPB}^{l}$.}
    \label{fig:jsf}
    \vspace{-1.0em}
\end{figure} 

Specifically, as shown in Figure~\ref{fig:jsf}, we take the $(l-1)$th layer feature $\mathbf{F}^{l-1}_{KPB} \in \mathbb{R}^{C_{l-1} \times H_{l-1} \times W_{l-1}}$ and the $l$th encoder feature $\mathbf{F}^{l}_E \in \mathbb{R}^{C_{l} \times H_{l} \times W_{l}}$ as input and extract attention vector $\alpha_{KPB}, \alpha_{E} \in \mathbb{R}^{C_{l}}$ by $3\times 3$ convolutions and MLP. Subsequently, the attention vector is divided into n groups with length m as $\Tilde{\alpha}_{KPB}, \Tilde{\alpha}_{E} \in \mathbb{R}^{n \times m}$. Thus, the channel-wise attention relation $ \mathbf{A} \in \mathbb{R}^{n \times n}$can be calculated by matrix product
\begin{equation}
    \mathbf{A} = \Tilde{\alpha}_{KPB} \odot \Tilde{\alpha}_{E}^{T}.
\end{equation}

After that, we calculate the selective factor $\overline{\alpha}_{KPB},\overline{\alpha}_{E} \in \mathbb{R}^{C_{l}}$ through $N \times N$ convolutions with splitting. Then, we obtain the selective attention weights $S^l \in \{ S_E^l, S_{KPB}^{l-1} \}$ for each features by $\alpha \in \{ \alpha_{KPB},\alpha_{E}\}$ and  $\overline{\alpha} \in \{ \overline{\alpha}_{KPB},\overline{\alpha}_{E} \}$, which can be formulated as
\begin{equation}
    S^l = \sigma ( \alpha + b \cdot FC(\overline{\alpha})), 
\end{equation}
where b is a learnable parameter, $\sigma$ is sigmoid function. Based on the attention weight vector, then the shallow and deep information are interacted as follows:
\begin{equation}
    \hat{\mathbf{F}}_{KPB}^{l} = S_{E}^{l} \cdot Conv(\mathbf{F}_{E}^{l}) + {\rm Upsample} ( S^{l-1}_{KPB} \cdot Conv(\mathbf{F}^{l-1}_{KPB})),
\end{equation}
where $\cdot$ denotes to element-wise multiplication.

\subsection{Harmony Kernel Modulation}
\label{sec:pixel_filtering}

The global-aware adaptive kernel obtained from the harmony kernel prediction branch is then utilized in harmony kernel modulation. In this section, we illustrate the regional harmony operation method kernel modulation in decoder layers, which converts the previous overall treatment of background and foreground into a local reference. As mentioned in Section~\ref{sec:HKP}, we apply multi-level kernel modulation in decoder layer to adapt scales and scenario variation problem. Figure~\ref{fig:main_model} shows our proposed kernel modulation in $l$th decoder layers, which could be formulated as
\begin{equation}
    \mathbf{\hat{F}}_{D}^l = \mathbf{F}_D^l \circledast \mathbf{K}^{l},
    \label{eq:filter}
\end{equation}
where $ \circledast$ denotes the harmony kernel modulation, $ \mathbf{F}_{D}^l \in \mathbb{R}^{C \times H \times W} $ is the deep feature extracted from the $l$th layer in decoder, and the $ \mathbf{\hat{F}_D}^l \in \mathbb{R}^{C \times H \times W} $ is its corresponding harmony kernel modulation result feature in the $l$th layer. The tensor $ \mathbf{K}^l \in \mathbb{R}^{C \times N^2 \times H \times W}$ represents the kernels with size of $N$ for harmony kernel modulation in the $l$th feature layer, which we obtain from KPB. For the kernel modulation in each pixel, we can expand the above equation as
\begin{equation}
    \mathbf{\hat{F}}_D^l[\mathbf{p}]=\sum_{\mathbf{q}\in \mathcal{N}_p} \mathbf{K}^l_p[\mathbf{p-q}]\mathbf{F}^l_D[\mathbf{q}],
    \label{eq:filterexpand}
\end{equation}
where $\mathbf{p}$ and $\mathbf{q}$ are the coordinates of pixels in the image, $\mathbf{K}^l_p$ is the kernel for filtering the element $\mathbf{p}$ of $\mathbf{F}_D^l$ via its surrounding elements, \ie,$\mathcal{N}_p$. As we illustrate in Eq.~\ref{eq:filter}, $\mathbf{K}^l$ contains all element-wise kernels,\ie, $\mathbf{K}^l_p \in \mathbb{R}^{C \times N \times N}$ for filtering operations. After the kernel modulation in decoder layers, we finally obtain the harmonization result. In this paper, we define the U-Net decoder layer as $\boldsymbol\phi_D(\cdot)$, then we can formulate our harmonization process with kernel modulation as 
$\hat{I}_h=\boldsymbol\phi_D^{L}(\cdots \boldsymbol\phi_D^{1}(\mathbf{F}^1_D \circledast \mathbf{K}^1)) \cdot M + (1-M) \cdot I_c$.

\subsection{Objective function}
In the training phase, we only employ foreground-normalized MSE loss as our objective function. Compared with normal MSE loss, it reduces the impact of copying background area:
\begin{equation}
    \mathcal{L}_{rec} = \dfrac{\sum\limits_{h,w} \left\Vert \hat{I} - I  \right\Vert_2^2}{\max \left \{ A_{min},\sum\limits_{h,w}M_{h,w} \right\} },
\end{equation}
where $A_{min}$ is a hyperparameter to keep the loss function stable as there might be some too small foreground objects. In this paper, we set $A_{min}=100$ as suggested in \cite{Sofiiuk_2021_WACV}.


\begin{table*}[!t]
  \centering
  \caption{Quantitative comparisons across four sub-datasets of iHarmony4~\cite{Cong_2020_CVPR}. $\uparrow$ indicates the higher the better, and $\downarrow$ indicates the lower the better. We compute fMSE for better reflection on harmonization tasks. Best results are in bold and the suboptimal results are in underline. }
  \resizebox{\linewidth}{!}{
    \begin{tabular}{cccccccccccccccc}
    \toprule[1.25pt]
    \multirow{2}[3]{*}{Method} & \multicolumn{3}{c}{HCOCO} & \multicolumn{3}{c}{HAdobe5k} & \multicolumn{3}{c}{HFlickr} & \multicolumn{3}{c}{Hday2night} & \multicolumn{3}{c}{ALL} \\
\cmidrule(r){2-4} \cmidrule(r){5-7} \cmidrule(r){8-10} \cmidrule(r){11-13} \cmidrule(r){14-16}
& PSNR↑ & MSE↓  & fMSE↓ & PSNR↑ & MSE↓  & fMSE↓ & PSNR↑ & MSE↓  & fMSE↓ & PSNR↑ & MSE↓  & fMSE↓ & PSNR↑ & MSE↓  & fMSE↓ \\
\midrule
    Composite & 33.94 & 69.37 & 996.59 & 28.16 & 345.54 & 2051.61 & 28.32 & 264.35 & 1574.37 & 34.01 & 109.65 & 1409.98 & 31.63 & 172.47 & 1376.42 \\
\midrule  DIH~\cite{Tsai_2017_CVPR}& 34.69 & 51.85 & 798.99 & 32.28 & 92.65 & 593.03 & 29.55 & 163.38 & 1099.13 & 34.62 & 82.34 & 1129.40 & 33.41 & 76.77 & 773.18 \\
   S$^2$AM~\cite{cun_2020_TIP} & 35.47 & 41.07 & 542.06 & 33.77 & 63.40  & 404.62 & 30.03 & 143.45 & 785.65 & 35.69 & 50.87 & 835.06 & 34.35 & 59.67 & 594.67 \\
    DoveNet~\cite{Cong_2020_CVPR}  & 35.83 & 36.72 & 551.01 & 34.34 & 52.32 & 380.39 & 30.21 & 133.14 & 827.03 & 35.27 & 51.95 & 1075.71 & 34.76 & 52.33 & 532.62 \\
    IIH~\cite{Guo_2021_CVPR} & 37.16 & 24.92 & 416.38 & 35.20  & 43.02 & 284.21 & 31.34 & 105.13 & 716.60 & 35.96 & 55.53 & 797.04 & 35.90  & 38.71 & 400.29 \\
    RAINNet~\cite{Ling_2021_CVPR} & 37.08 & 29.52 & 501.17 & 36.22 & 43.35 & 317.35 & 31.64 & 110.59 & 688.40 & 34.83 & 57.40  & 916.48 & 36.12 & 40.29 & 469.60 \\
    iDIH-HRNet~\cite{Sofiiuk_2021_WACV}  & 39.16 & 16.48 & 266.19 & 38.08 & 21.88 & 173.96 & 33.13 & 69.67 & 443.65 & 37.72 & \underline{40.59} & 590.97 & 38.19 & 24.44 & 264.96 \\
    D-HT~\cite{Guo_2021_ICCV} & 38.76 & 16.89 & 299.30 & 36.88 & 38.53 & 265.11 & 33.13 & 74.51 & 515.45 & 37.10  & 53.01 & 704.42 & 37.55 & 30.30  & 320.78 \\
    Harmonizer~\cite{Harmonizer} & 38.77 & 17.34 & 298.42 & 37.64 & 21.89 & 170.05 & 33.63 & 64.81 & 434.06 & 37.56 & {\bf 33.14} & {\bf 542.07} & 37.84 & 24.26 & 280.51 \\ 
    DCCF~\cite{xue2022dccf} & 39.72 & 14.55 & 267.79 & 38.24 & \underline{20.20} & 171.01 & 33.72 & 66.20 & 440.84 & \underline{38.18} & 51.40 & 629.67 & 38.60 & 22.64& 265.41 \\ 
    SCS-Co~\cite{Hang_2022_CVPR} & \underline{39.88} & \underline{13.58} & \underline{245.54} & \underline{38.29} & 21.01 & \underline{165.48} & \underline{34.22} & {\bf 55.83} & \underline{393.72} & 37.83 & 41.75 & 606.80 & \underline{38.75} & \underline{21.33} & \underline{248.86} \\
    \midrule
    Ours & {\bf 40.32} & {\bf 12.95} & {\bf 222.31} & {\bf 39.97} & {\bf 17.84} & {\bf 138.22} & {\bf 34.45} & \underline{57.58} & {\bf 372.90} & {\bf 38.47} & 42.76 &  \underline{546.06} & {\bf 39.53} & {\bf 19.90} & {\bf 220.44} \\
    \bottomrule[1.25pt]
    \end{tabular}%
    }

  \label{tab:sotas}%
  \vspace{-1.0em}
\end{table*}

\begin{table}[!t]
  \centering
  \caption{Quantitative comparisons on different ratios of foreground based on iHarmony4 by MSE and fMSE metrics. The best results are in bold and the suboptimal results are in underline.}
  \resizebox{\linewidth}{!}{
    \begin{tabular}{ccrrrrrr}
    \toprule[1.25pt]
    \multicolumn{1}{c}{\multirow{2}[2]{*}{Method}} & \multicolumn{1}{c}{\multirow{2}[2]{*}{Venue}} & \multicolumn{2}{c}{0\%$ \sim$5\%} & \multicolumn{2}{c}{5\%$\sim$15\%} & \multicolumn{2}{c}{15\%$\sim$100\%} \\
    \cmidrule(r){3-4} \cmidrule(r){5-6} \cmidrule(r) {7-8} 
          &       & \multicolumn{1}{l}{MSE$\downarrow$} & \multicolumn{1}{l}{fMSE$\downarrow$} & \multicolumn{1}{l}{MSE$\downarrow$} & \multicolumn{1}{l}{fMSE$\downarrow$} & \multicolumn{1}{l}{MSE$\downarrow$} & \multicolumn{1}{l}{fMSE$\downarrow$} \\ 
    \midrule
    DIH   & CVPR'17 & 18.92      & 799.17      &    64.23   & 725.86      & 228.86      & 768.89      \\
    S$^2$AM & TIP'20 & 15.09      & 623.11      & 48.33      & 540.54      &    117.62   & 592.83      \\
    DoveNet & CVPR'20 & 14.03      &    591.88   & 44.90      & 504.42      & 152.07      & 505.82      \\
    RainNet & CVPR'21 & 11.66      &    550.38   & 32.05      & 378.69      & 117.41      & 389.80      \\
    iS$^2$AM & WACV'21 &\underline{6.73}       & \underline{294.76}       & \underline{18.03}       & \underline{204.69}      & \underline{63.02}      &  \underline{207.82}       \\
    \midrule
    GKNet   & Ours  &  {\bf 5.36}     & {\bf 244.06}      &  {\bf 17.46}     & {\bf 200.34}      &    {\bf 57.31}   & {\bf 188.75}      \\
    \bottomrule[1.25pt]
    \end{tabular}}

  \label{tab:ratios}%
  \vspace{-1.0em}
\end{table}%

\section{Experiments}

\subsection{Implementation Details}

We conduct the image harmonization experiment at resolution $256 \times 256$ on the benchmark dataset iHarmony4~\cite{Cong_2020_CVPR}. The initial learning rate is set to $10^{-4}$, and the models are trained for 120 epochs with a batch size of 16 on four 2080Ti GPUs. For optimizer, we adopt an Adam optimizer with $\beta_{1}=0.9,\beta_{2}=0.999$ and $\epsilon=10^{-8}$. It takes about two days for training. Our proposed model is implemented by PyTorch, and more detailed network architectures could be found in the supplementary file.

\subsection{Datasets and Metrics}

\noindent \textbf{Datasets.}
To evaluate our proposed method for image harmonization, we conduct our experiments on the benchmark dataset iHarmony4~\cite{Cong_2020_CVPR}, which consists of 4 sub-datasets: HCOCO, HAdobe5K, HFlickr, and Hday2night, including 73147 pairs of synthesized composite images with their corresponding foreground mask and ground truth image. 

\noindent \textbf{Evaluation Metrics.}
Following the standard setups in image harmonization, we use the peak signal-to-noise ratio (PSNR) and Mean Squared Error (MSE) as evaluation metrics. Furthermore, it is more accurate to only calculate the difference in the foreground region with the metric foreground MSE (fMSE) introduced by~\cite{Sofiiuk_2021_WACV}. The metric MSE is calculated for the whole image, while there is no changes for pixels in the background region in harmonization task. Without considering the foreground ratio, the average MSE and PSNR results of the dataset will be more responsive to the performance of large-scale targets. In this paper, we argue that fMSE is more suitable for harmonization task.

\begin{figure*}[h]
\captionsetup[subfloat]{labelformat=empty}
\centering
\subfloat[Mask]{
    \begin{minipage}[b]{0.13\linewidth}
    \includegraphics[width=1\linewidth]{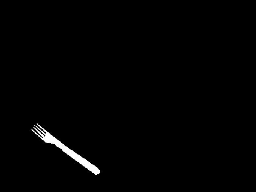}\vspace{1pt}
    \includegraphics[width=1\linewidth]{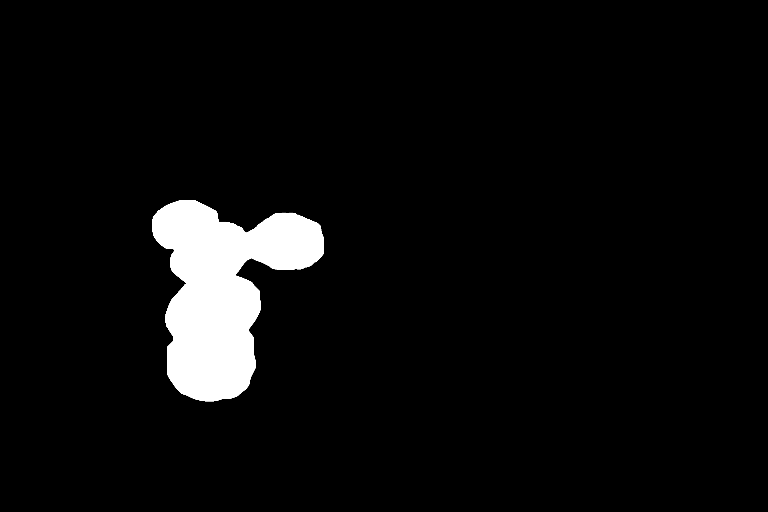}\vspace{1pt}
    \includegraphics[width=1\linewidth]{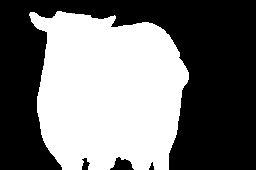}\vspace{1pt}
    \includegraphics[width=1\linewidth]{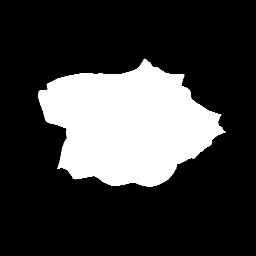}\vspace{1pt}
    \end{minipage}}
\subfloat[Input]{
    \begin{minipage}[b]{0.13\linewidth}
    \includegraphics[width=1\linewidth]{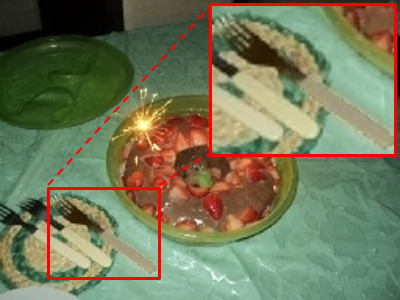}\vspace{1pt}
    \includegraphics[width=1\linewidth]{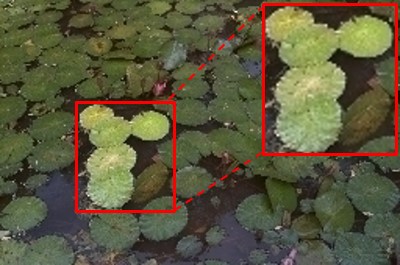}\vspace{1pt}
    \includegraphics[width=1\linewidth]{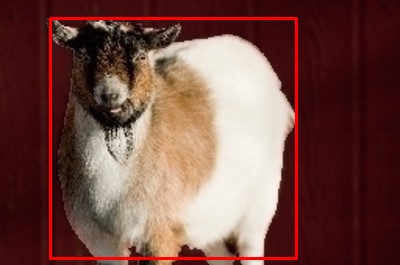}\vspace{1pt}
    \includegraphics[width=1\linewidth]{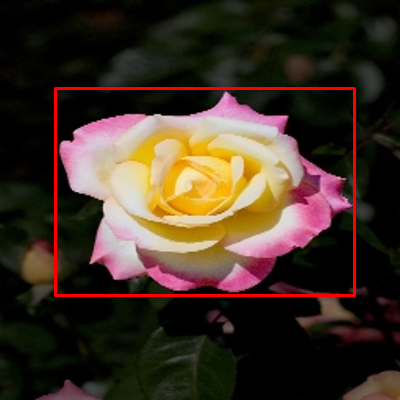}\vspace{1pt}
    \end{minipage}}
\subfloat[RainNet~\cite{Ling_2021_CVPR}]{
    \begin{minipage}[b]{0.13\linewidth}
    \includegraphics[width=1\linewidth]{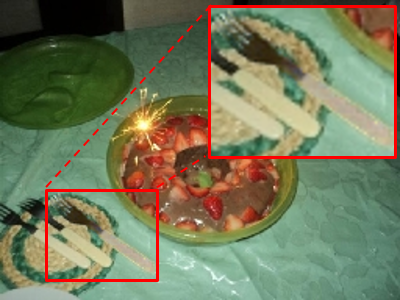}\vspace{1pt}
    \includegraphics[width=1\linewidth]{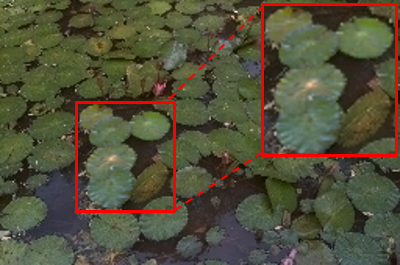}\vspace{1pt}
    \includegraphics[width=1\linewidth]{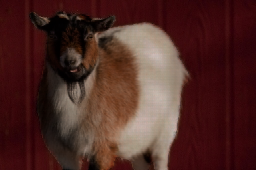}\vspace{1pt}
    \includegraphics[width=1\linewidth]{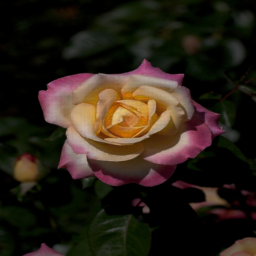}\vspace{1pt}
    \end{minipage}}
\subfloat[D-HT~\cite{Guo_2021_ICCV}]{
    \begin{minipage}[b]{0.13\linewidth}
    \includegraphics[width=1\linewidth]{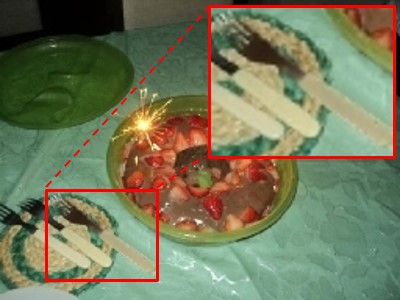}\vspace{1pt}
    \includegraphics[width=1\linewidth]{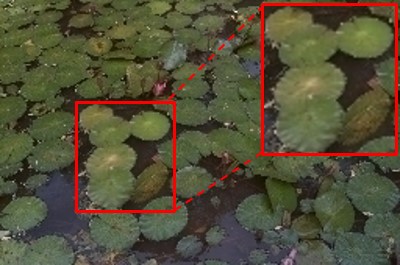}\vspace{1pt}
    \includegraphics[width=1\linewidth]{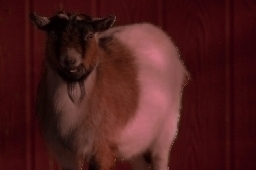}\vspace{1pt}
    \includegraphics[width=1\linewidth]{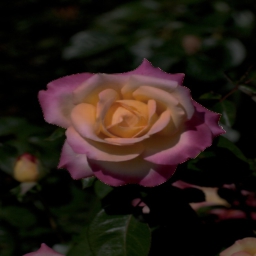}\vspace{1pt}
    \end{minipage}}
\subfloat[Harmonizer~\cite{Harmonizer}]{
    \begin{minipage}[b]{0.13\linewidth}
    \includegraphics[width=1\linewidth]{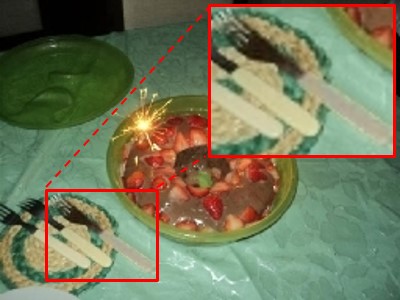}\vspace{1pt}
    \includegraphics[width=1\linewidth]{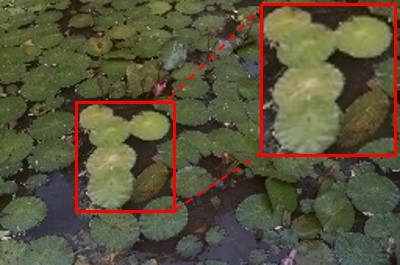}\vspace{1pt}
    \includegraphics[width=1\linewidth]{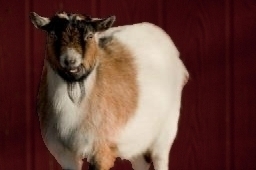}\vspace{1pt}
    \includegraphics[width=1\linewidth]{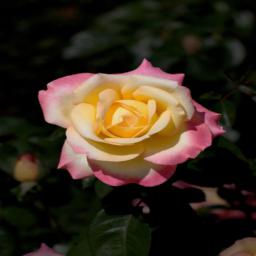}\vspace{1pt}
    \end{minipage}}
\subfloat[Ours]{
    \begin{minipage}[b]{0.13\linewidth}
    \includegraphics[width=1\linewidth]{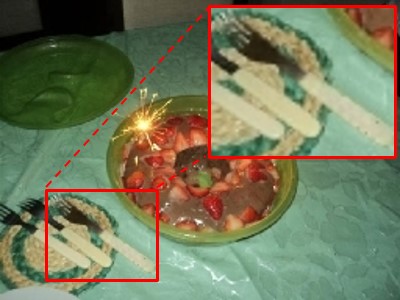}\vspace{1pt}
    \includegraphics[width=1\linewidth]{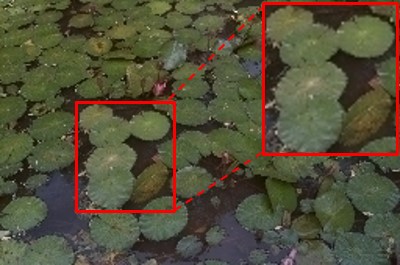}\vspace{1pt}
    \includegraphics[width=1\linewidth]{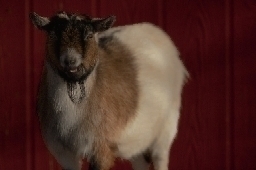}\vspace{1pt}
    \includegraphics[width=1\linewidth]{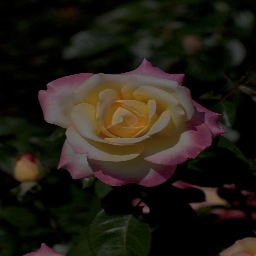}\vspace{1pt}
    \end{minipage}}
\subfloat[Real(GT)]{
    \begin{minipage}[b]{0.13\linewidth}
    \includegraphics[width=1\linewidth]{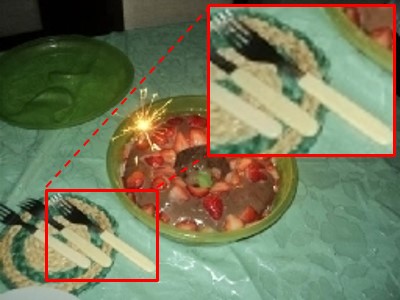}\vspace{1pt}
    \includegraphics[width=1\linewidth]{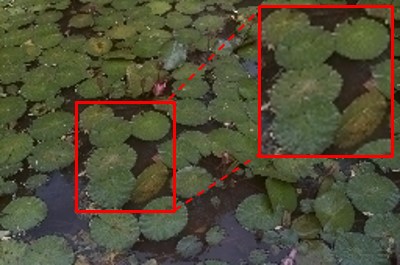}\vspace{1pt}
    \includegraphics[width=1\linewidth]{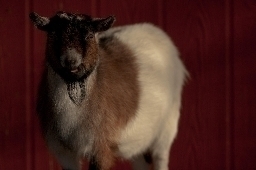}\vspace{1pt}
    \includegraphics[width=1\linewidth]{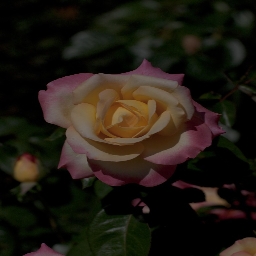}\vspace{1pt}
    \end{minipage}}
    \caption{{\bf Qualitative comparisons with SOTA methods on iHarmony4~\cite{Cong_2020_CVPR}.} Mask in column one and Red box in Input represents the foreground. \textbf{Case 1:} The background color shows spatial variation, while only our method captures practical color reference by predicted kernel and local harmony modulation. \textbf{Case 2:} The harmony result obtained by our method is better in the detailed structure like duckweed center due to target harmony kernels for each foreground part. \textbf{Case 3 \& 4:} For large foregrounds, our approach could also achieve better results, which preserves more original details and is closer to ground truth.}
    \label{fig:sota_vis}
\end{figure*}

\begin{figure*}[t]
    \centering
    \captionsetup[subfloat]{labelformat=empty}
    \subfloat[Mask]{
    \begin{minipage}[b]{0.15\linewidth}
    \includegraphics[width=1\linewidth]{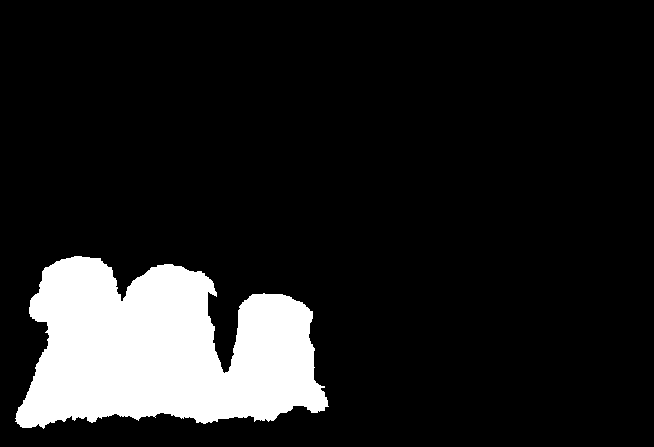}\vspace{1pt}
    \includegraphics[width=1\linewidth]{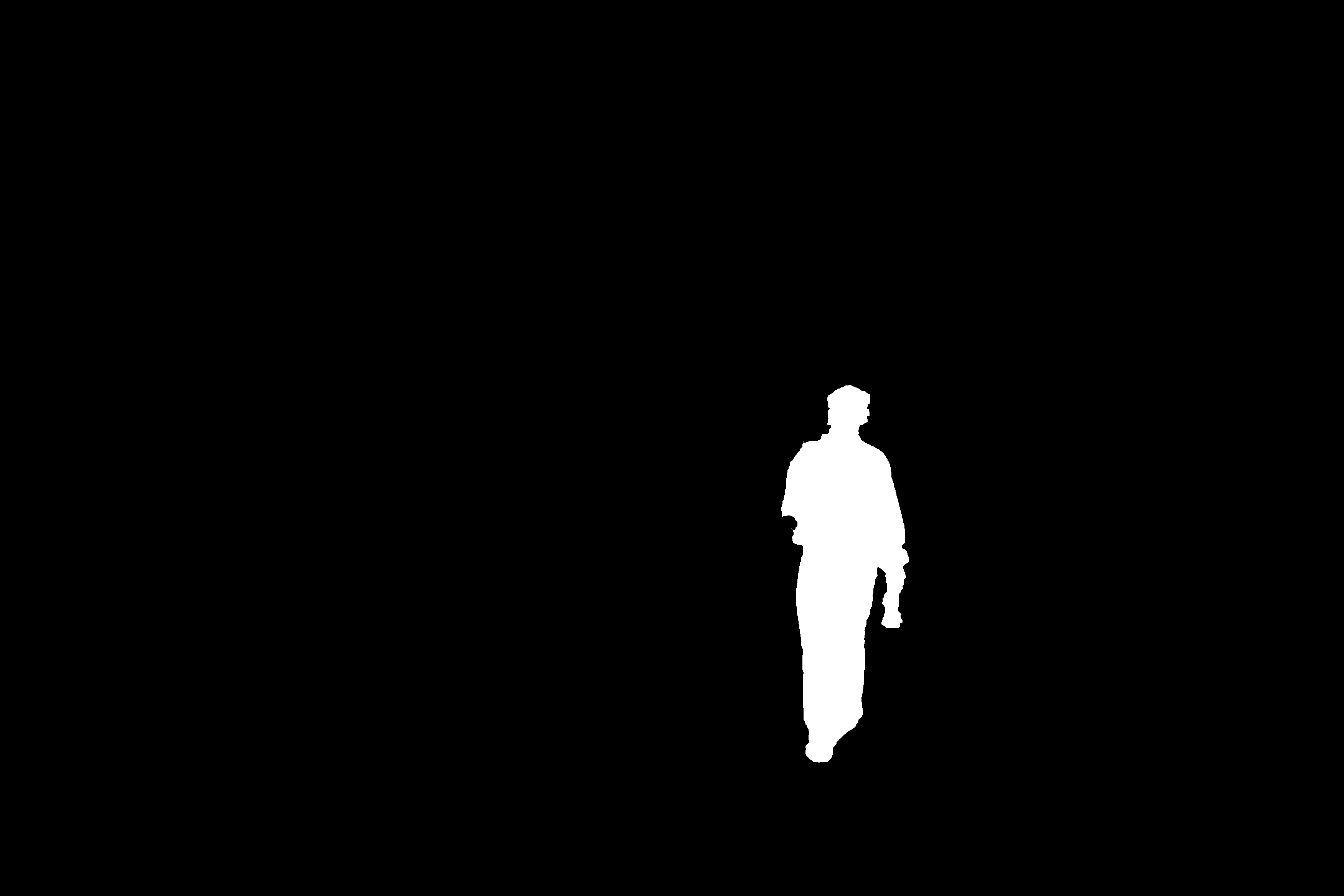}\vspace{1pt}
    \end{minipage}}
\subfloat[Input]{
    \begin{minipage}[b]{0.15\linewidth}
    \includegraphics[width=1\linewidth]{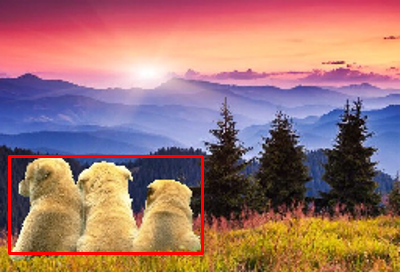}\vspace{1pt}
    \includegraphics[width=1\linewidth]{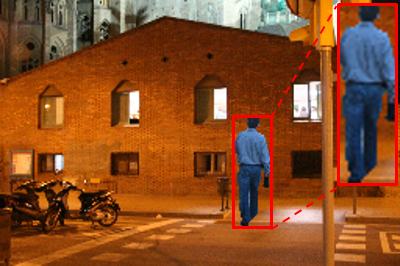}\vspace{1pt}
    \end{minipage}}
\subfloat[RainNet~\cite{Ling_2021_CVPR}]{
    \begin{minipage}[b]{0.15\linewidth}
    \includegraphics[width=1\linewidth]{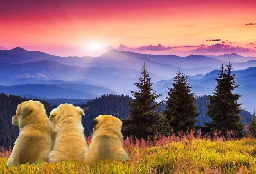}\vspace{1pt}
    \includegraphics[width=1\linewidth]{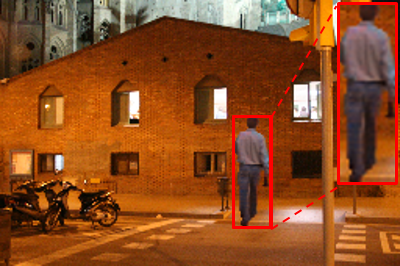}\vspace{1pt}
    \end{minipage}}
\subfloat[iDIH-HRNet~\cite{Sofiiuk_2021_WACV}]{
    \begin{minipage}[b]{0.15\linewidth}
    \includegraphics[width=1\linewidth]{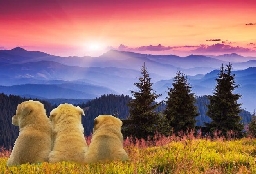}\vspace{1pt}
    \includegraphics[width=1\linewidth]{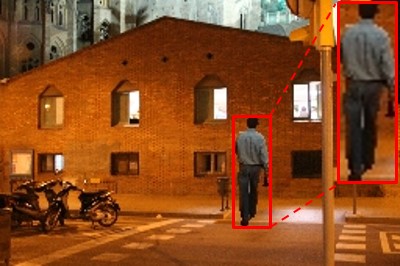}\vspace{1pt}
    \end{minipage}}
\subfloat[DoveNet~\cite{Cong_2020_CVPR}]{
    \begin{minipage}[b]{0.15\linewidth}
    \includegraphics[width=1\linewidth]{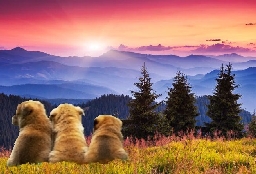}\vspace{1pt}
    \includegraphics[width=1\linewidth]{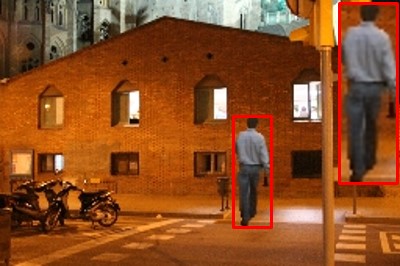}\vspace{1pt}
    \end{minipage}}
\subfloat[Ours]{
    \begin{minipage}[b]{0.15\linewidth}
    \includegraphics[width=1\linewidth]{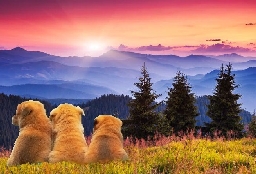}\vspace{1pt}
    \includegraphics[width=1\linewidth]{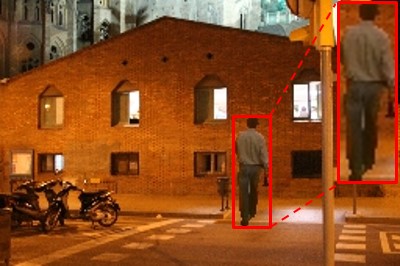}\vspace{1pt}
    \end{minipage}}

    \caption{{\bf Qualitative comparisons with SOTA methods on real composite images}~\cite{Tsai_2017_CVPR}. Mask in column one and Red box in Input represents the foregrounds. \textbf{Case 1:} Comparing the results on the dogs, our method achieves more natural results over the sunset background.  \textbf{Case 2:} Our result dims the foreground person based on background dim light, which achieves a more natural effect. }
    \label{fig:real_vis}
    \vspace{-1.0em}
\end{figure*}

\subsection{Comparison with SOTAs}

\paragraph{Quantitative comparison.}
As results shown in Table~\ref{tab:sotas}, we compare our method with other state-of-the-art image harmonization methods on iHarmony4~\cite{Cong_2020_CVPR}. Following~\cite{Cong_2020_CVPR,Sofiiuk_2021_WACV}, we also evaluate the model performance on different ratios of foreground by splitting the test images into three groups,\ie, {0\% $\sim$5\%}, {5\%$\sim$15\%}, and {15\%$\sim$100\%}. We provide these results in Table~\ref{tab:ratios}. Observing the quantitative experiment results above, we can summarize the following conclusions: {\bf 1)} Our method achieves SOTA results of all evaluation metrics on average iharmony4 datasets. More specifically, our method achieves 0.78dB$\uparrow$ improvement in PSNR, 1.43$\downarrow$ in MSE, and 28.42$\downarrow$ in fMSE compared with suboptimal methods. {\bf 2)} Our method obtains the best fMSE scores on all sub-datasets, meaning that the foreground regions generated by our method are more natural and closer to real images. {\bf 3)} As shown in Table~\ref{tab:ratios}, our model performs well on each foreground ratio, especially on {0\%$\sim$5\%}, which indicates that our method has a strong ability to handle variable foreground scales. It also proves that our proposed global-aware adaptive kernel could process global-to-local impact with excellence.

\noindent {\bf Qualitative comparison.}
We further provide qualitative results on iHarmony4 datasets in Figure~\ref{fig:sota_vis}. It could be observed that our GKNet generates more visually coherent images of the foreground and background than composited images, which are also closer to ground truth. 
{\bf 1)} The first two rows of examples show that our method can effectively handle the small object harmonization problem in complex scenes with proximity prior consideration.
{\bf 2)} The last two row examples show that our regional modulation with multi-level structure performs well in large foreground cases, which handles a wide range of enormous contrast. For more detailed descriptions, please refer to the caption in Figure~\ref{fig:sota_vis}.
In contrast, our method presents more photorealistic results. More visual comparison results on iHarmony4 datasets could be seen in supplementary materials.

\noindent{\bf Comparisions on Real Datasets.}
Experiments results on real composition datasets~\cite{Tsai_2017_CVPR} can be seen in Figure~\ref{fig:real_vis}. For real composition cases, evaluation metrics are impossible to calculate as there are no ground truth images. Hence, we only show qualitative results and human study results here. More visual comparison results on real datasets could be seen in supplementary materials. 

\begin{table}[ht]
  \centering
    \caption{B-T scores comparison on real composite images.}
    \resizebox{0.95\linewidth}{!}{
    \begin{tabular}{lccccc}
    \toprule[1.25pt]
    \multicolumn{1}{c}{Method} & \multicolumn{1}{c}{Composite} & \multicolumn{1}{c}{DoveNet\cite{Cong_2020_CVPR}} & \multicolumn{1}{c}{RainNet\cite{Ling_2021_CVPR}} & \multicolumn{1}{c}{iDIH-HRNet\cite{Sofiiuk_2021_WACV}} & \multicolumn{1}{c}{Ours} \\
    \midrule
    B-T Score $\uparrow$ & 0.416      & 0.686      &    0.972   &  1.532     & {\bf 1.944} \\
    \bottomrule[1.25pt]
    \end{tabular}%
    }
  \label{tab:bt_score}%
  \vspace{-1.0em}
\end{table}%

Following~\cite{Cong_2020_CVPR,cong_2021_ICME,Guo_2021_CVPR,Hang_2022_CVPR}, we conduct our human study by inviting 50 volunteers to compare 24750 pairwise results. The pairwise results are obtained from 99 real composited images, with 25 results for each pair of different methods on average. We also use the Bradley-Terry model (B-T model) to calculate the global ranking score. Our method achieves the best results as shown in Table~\ref{tab:bt_score}.

\subsection{Ablation Studies}

\begin{figure}[t]
    \centering
    \captionsetup[subfloat]{labelformat=empty}
    \resizebox{\linewidth}{!}{
    \subfloat[Input]{
    \begin{minipage}[b]{0.2\linewidth}
    \includegraphics[width=1\linewidth]{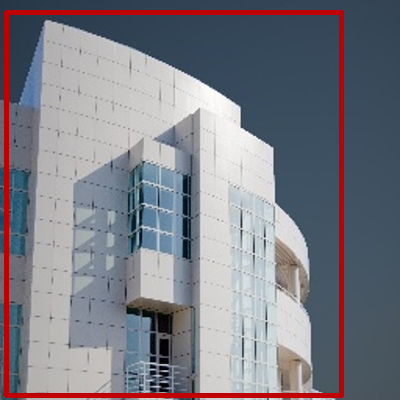}\vspace{1pt}
    \includegraphics[width=1\linewidth]{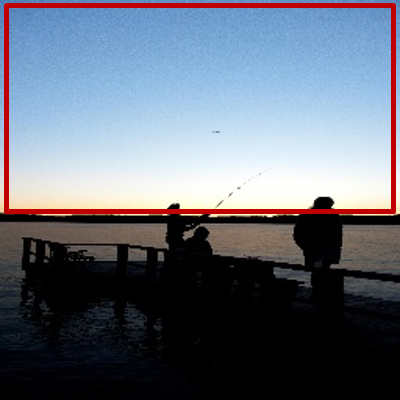}\vspace{1pt}
    \end{minipage}}
\subfloat[GT]{
    \begin{minipage}[b]{0.2\linewidth}
    \includegraphics[width=1\linewidth]{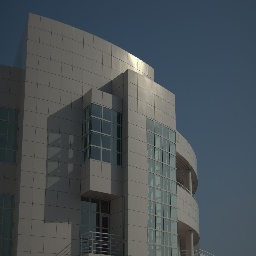}\vspace{1pt}
    \includegraphics[width=1\linewidth]{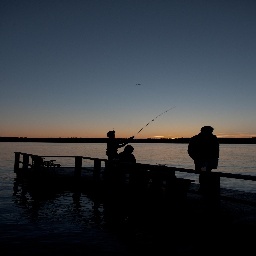}\vspace{1pt}
    \end{minipage}}
\subfloat[Baseline]{
    \begin{minipage}[b]{0.2\linewidth}
    \includegraphics[width=1\linewidth]{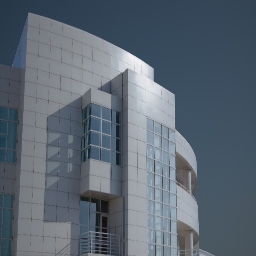}\vspace{1pt}
    \includegraphics[width=1\linewidth]{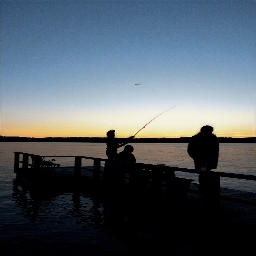}\vspace{1pt}
    \end{minipage}}
\subfloat[KPB]{
    \begin{minipage}[b]{0.2\linewidth}
    \includegraphics[width=1\linewidth]{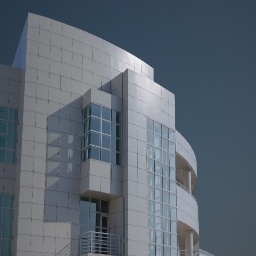}\vspace{1pt}
    \includegraphics[width=1\linewidth]{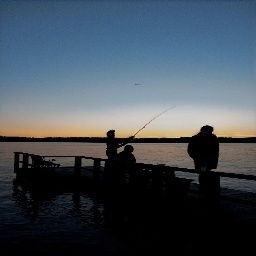}\vspace{1pt}
    \end{minipage}}
\subfloat[LRE]{
    \begin{minipage}[b]{0.2\linewidth}
    \includegraphics[width=1\linewidth]{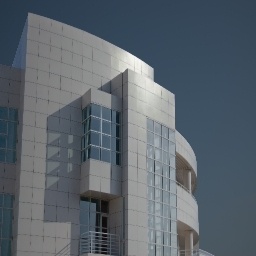}\vspace{1pt}
    \includegraphics[width=1\linewidth]{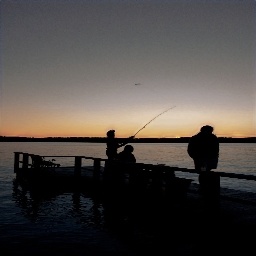}\vspace{1pt}
    \end{minipage}}
\subfloat[Full Model]{
    \begin{minipage}[b]{0.2\linewidth}
    \includegraphics[width=1\linewidth]{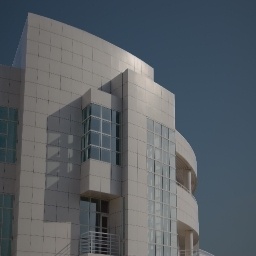}\vspace{1pt}
    \includegraphics[width=1\linewidth]{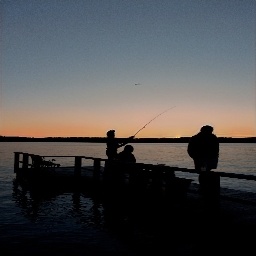}\vspace{1pt}
    \end{minipage}}}
    \caption{Qualitative ablation study of our approach. Red boxes in input image mark foreground.}
    \label{fig:qualitative_as}
    \vspace{-0.5em}
\end{figure}

\noindent \textbf{Effectiveness of network components.} We further conduct quantitative experiments to verify the effectiveness of components in GKNet. Note that modules in GKNet have dependencies, we can only show gradually added ablation studies.
As the results in Table~\ref{tab:components_as}, our full model obtains the highest performance on all metrics when KPB, LRE, and SCF work together. Table~\ref{tab:components_as} also illustrates the effectiveness of each component.
Moreover, to further illustrate our global-local interaction method for harmonization, we show qualitative results in Figure~\ref{fig:qualitative_as}. Compared with the baseline, the model with only local KPB outputs stronger local modeling results but loses global view. After introducing LRE, global awareness of the background is enhanced.But it is evident that color is overcorrected in the second row. After adding SCF, we effectively correct the deviation by selecting relevant references in background, and the final result is closer to real image.

\begin{table}[t]
  \centering
    \caption{Quantitative ablation study of our approach with different components on iHarmony4\cite{Cong_2020_CVPR}}
    \resizebox{0.95\linewidth}{!}{
    \begin{tabular}{ccccccc}
    \toprule[1.25pt]
    Baseline & KPB   & LRE   & SCF  & \multicolumn{1}{l}{MSE $\downarrow$} & \multicolumn{1}{l}{PSNR $\uparrow$} & \multicolumn{1}{l}{fMSE $\downarrow$ } \\
    \midrule
    \ding{51} & \ding{55} & \ding{55}  & \ding{55}     &   27.27    & 37.83      & 280.56 \\
    \ding{51} & \ding{51} & \ding{55}  & \ding{55}    &  
    21.42     & 38.68      & 235.72 \\
    \ding{51} & \ding{51} & \ding{51} & \ding{55}      &    20.50  & 39.30      & 229.63 \\
    \ding{51} & \ding{51} & \ding{51} & \ding{51} & 19.90      &    39.53   & 220.44  \\
    \bottomrule[1.25pt]
    \end{tabular}%
    }
  \label{tab:components_as}%
\end{table}%

\begin{figure}[h]
	\centering
	\resizebox{\linewidth}{!}{
	\subfloat{
        \rotatebox{90}{\scriptsize{~Harmonized}}
		\begin{minipage}[t]{0.16\linewidth}
			\centering
			\includegraphics[width=1\linewidth]{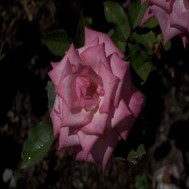}
		\end{minipage}
		\begin{minipage}[t]{0.16\linewidth}
			\centering
			\includegraphics[width=1\linewidth]{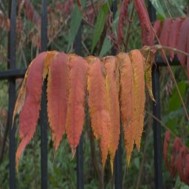}
		\end{minipage}
		\begin{minipage}[t]{0.16\linewidth}
			\centering
			\includegraphics[width=1\linewidth]{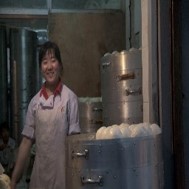}
		\end{minipage}
		\begin{minipage}[t]{0.16\linewidth}
			\centering
			\includegraphics[width=1\linewidth]{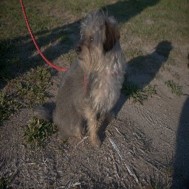}
		\end{minipage}
		\begin{minipage}[t]{0.16\linewidth}
			\centering
			\includegraphics[width=1\linewidth]{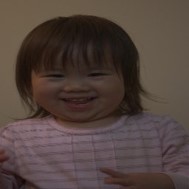}
		\end{minipage}
		\begin{minipage}[t]{0.16\linewidth}
			\centering
			\includegraphics[width=1\linewidth]{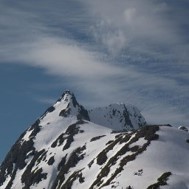}
		\end{minipage}
	}
	}
	\resizebox{\linewidth}{!}{
    \subfloat{
		\rotatebox{90}{\scriptsize{~~~~~~~Mask}}
		\begin{minipage}[t]{0.16\linewidth}
			\centering
			\includegraphics[width=1\linewidth]{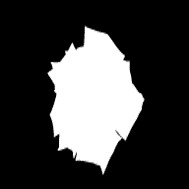}
		\end{minipage}
		\begin{minipage}[t]{0.16\linewidth}
			\centering
			\includegraphics[width=1\linewidth]{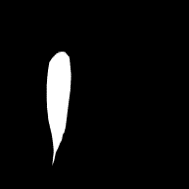}
		\end{minipage}
		\begin{minipage}[t]{0.16\linewidth}
			\centering
\includegraphics[width=1\linewidth]{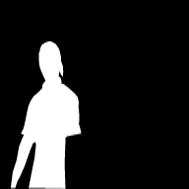}
		\end{minipage}
		\begin{minipage}[t]{0.16\linewidth}
			\centering
			\includegraphics[width=1\linewidth]{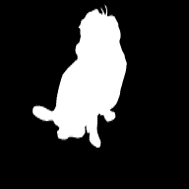}
		\end{minipage}
		\begin{minipage}[t]{0.16\linewidth}
			\centering
			\includegraphics[width=1\linewidth]{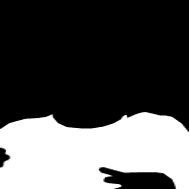}
		\end{minipage}
		\begin{minipage}[t]{0.16\linewidth}
			\centering
			\includegraphics[width=1\linewidth]{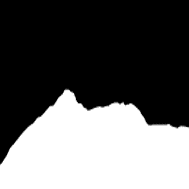}
		\end{minipage}
	}}
	\resizebox{\linewidth}{!}{
	\subfloat{
		\rotatebox{90}{\scriptsize{~~~~~~Kernel}}
		\begin{minipage}[t]{0.16\linewidth}
			\centering
			\includegraphics[width=1\linewidth]{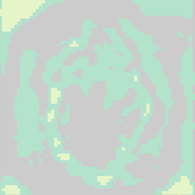}
		\end{minipage}
		\begin{minipage}[t]{0.16\linewidth}
			\centering
            \includegraphics[width=1\linewidth]{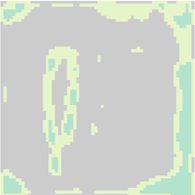}
		\end{minipage}
		\begin{minipage}[t]{0.16\linewidth}
			\centering
            \includegraphics[width=1\linewidth]{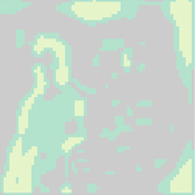}
		\end{minipage}
		\begin{minipage}[t]{0.16\linewidth}
			\centering
			\includegraphics[width=1\linewidth]{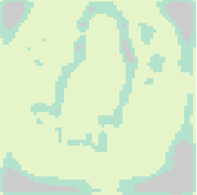}
		\end{minipage}
		\begin{minipage}[t]{0.16\linewidth}
			\centering
			\includegraphics[width=1\linewidth]{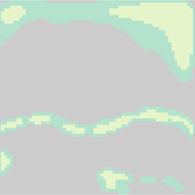}
		\end{minipage}
		\begin{minipage}[t]{0.16\linewidth}
			\centering
			\includegraphics[width=1\linewidth]{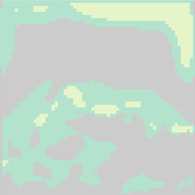}
		\end{minipage}
	}}
	\caption{The cluster visual results of adaptive harmony kernels. \eg, in flower case, structure information like petal margin is displayed, which proves that kernels are predicted to adapt local harmonization specifically. }
	\label{fig:kernel}
    \vspace{-0.5em}
\end{figure}

\noindent{\bf Interpretability of Global-aware Harmony Kernel.}
To further illustrate the effectiveness of the adaptive harmony kernels, we cluster the per-pixel adaptive harmony kernels predicted from KPB by K-means. As shown in Figure~\ref{fig:kernel}, the clusters show strong spatial structure, which indicates that \emph{our predicted dynamic kernel can make the structural adjustment to harmonize the foreground}. Moreover, in some cases, the change of kernel classes is related to the object mask. This exhibits that our harmony kernels are predicted dynamically to deal with visual inconsistency in different spatial locations (\eg fore-/-background or edges).

\begin{figure}[t]
    \centering
    \resizebox{\linewidth}{!}{
    \subfloat[Mask]{
    \begin{minipage}[t]{0.16\linewidth}
        \centering
        \includegraphics[width=1\linewidth]{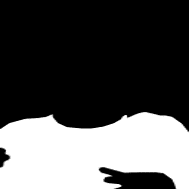}
        \includegraphics[width=1\linewidth]{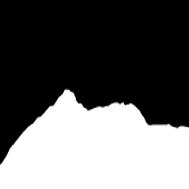}
    \end{minipage}}\hspace{-1mm}
    \subfloat[Comp]{
        \begin{minipage}[t]{0.16\linewidth}
        \centering
        \includegraphics[width=1\linewidth]{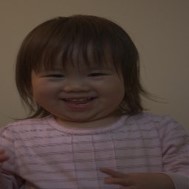}
        \includegraphics[width=1\linewidth]{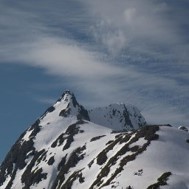}
    \end{minipage}}\hspace{-1.5mm}
    \subfloat[Feature map with channel attention weight]{
        \begin{minipage}[t]{0.16\linewidth}
        \centering
        \includegraphics[width=1\linewidth]{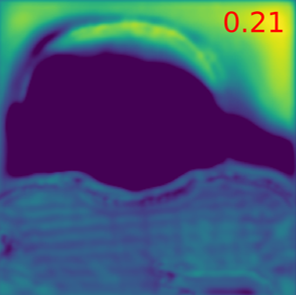}
        \includegraphics[width=1\linewidth]{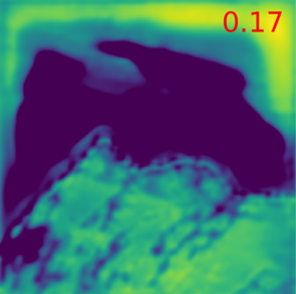}
    \end{minipage}
        \begin{minipage}[t]{0.16\linewidth}
        \centering
        \includegraphics[width=1\linewidth]{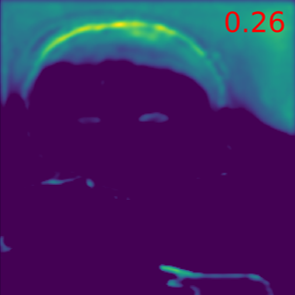}
        \includegraphics[width=1\linewidth]{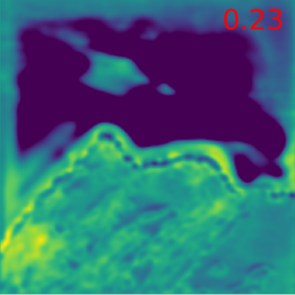}
    \end{minipage}
        \begin{minipage}[t]{0.16\linewidth}
        \centering
        \includegraphics[width=1\linewidth]{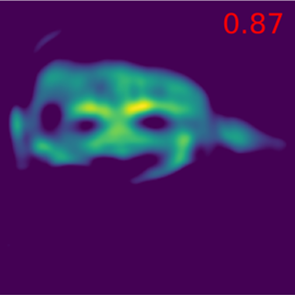}
        \includegraphics[width=1\linewidth]{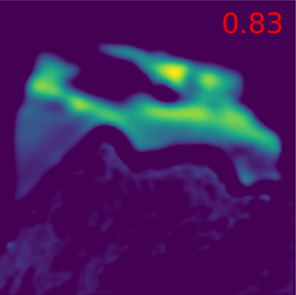}
    \end{minipage}
        \begin{minipage}[t]{0.16\linewidth}
        \centering
        \includegraphics[width=1\linewidth]{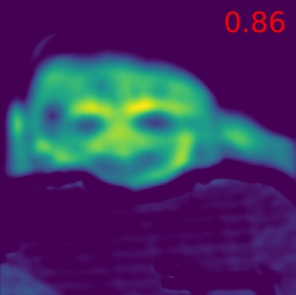}
        \includegraphics[width=1\linewidth]{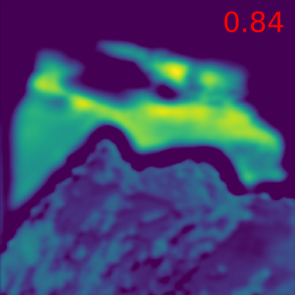}
    \end{minipage}
    }}
    \caption{Feature maps in SCF. Red number in upper right corner of feature map represents channel-wise attention weights.}
    \label{fig:channel}
    \vspace{-1.0em}
\end{figure}

\begin{figure}[t]
    \centering
    \includegraphics[width=0.9\linewidth]{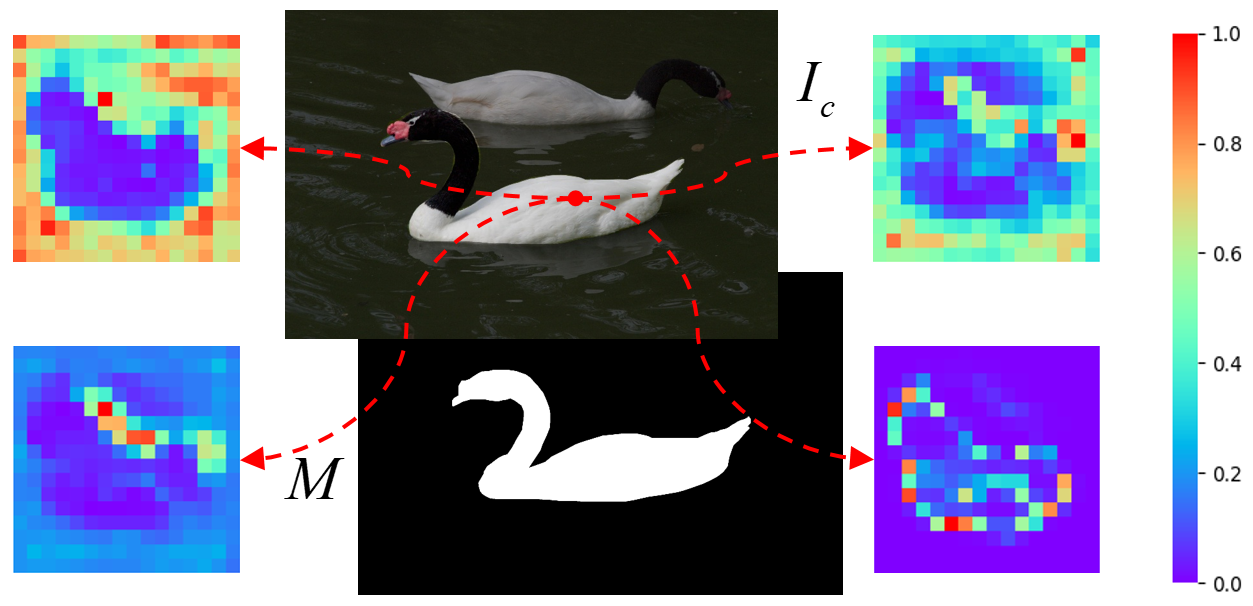}
    \caption{ Cross-attention from example point (10,10). We show four attention maps for different heads, which proves LRE can match model related long-distance references.}
    \label{fig:transformer}
    \vspace{-1.0em}
\end{figure}

\noindent{\bf Interpretability of LRE and SCF Module.}
We visualize the attention mechanism in LRE and SCF for interpreting global-local interaction. In Figure~\ref{fig:channel}, we visualize feature maps in decoder layers to illustrate our channel-wise attention module SCF. Visual feature maps with attention weights show that harmony kernels are predicted with more attention on related background area and less attention on irrelevant background area or foreground. In Figure~\ref{fig:transformer}, we visualize the attention maps of LRE, focusing on an example point in foreground. As the long-distance information in predicted harmony kernels is brought by LRE, the visualized attention maps in different heads indicate two points: {\bf 1)} The kernels for local operation have a global perceptive field with long-distance information. {\bf 2)} Different heads pay attention to different reference parts, \ie relevant background reference, foreground content, overall tone, etc.

\section{Conclusion}

This paper proposes an effective network GKNet to learn global-aware harmony kernels for image harmonization, including harmony kernel prediction and harmony kernel modulation branches. For harmony kernel prediction, we propose LRE to extract long-term references and KPB to predict global-aware kernels. To better fuse long-term context, we design SCF to select relevant references. For harmony kernel modulation, we employ the predicted kernels for harmonization with location awareness. Extensive experiments demonstrate that our proposed algorithm outperforms the state-of-the-art algorithms on the iHarmony4 dataset and real image composition datasets.

\noindent{\bf Acknowledgements.} This work was supported by a Grant from The National Natural Science Foundation of China (No. 2021YFB2012300).

{\small
\bibliographystyle{ieee_fullname}
\bibliography{egbib}
}

\end{document}